\documentclass[sigconf]{acmart}

\usepackage{booktabs} 

\usepackage{tabularx}
\usepackage{multirow}
\usepackage{subfigure}

\usepackage{color}
\usepackage[ruled]{algorithm2e} 

\SetAlFnt{\small}
\SetAlCapFnt{\small}
\SetAlCapNameFnt{\small}
\SetAlCapHSkip{0pt}
\IncMargin{-\parindent}

\usepackage{arydshln}

\usepackage{subfigure}
\usepackage{amsmath, amssymb}
\usepackage{algorithmic}
\usepackage{tabularx}
\usepackage{color}
\SetKwInput{kwInitStep}{Initialization Step}
\SetKwInput{kwGibbsStep}{Gibbs Sampling Step}

\usepackage{amsmath}

\setcopyright{acmcopyright}

\copyrightyear{2020}
\acmYear{2020}
\setcopyright{iw3c2w3}
\acmConference[WWW '20]{Proceedings of The Web Conference 2020}{April 20--24, 2020}{Taipei, Taiwan}
\acmBooktitle{Proceedings of The Web Conference 2020 (WWW '20), April 20--24, 2020, Taipei, Taiwan}
\acmPrice{}
\acmDOI{10.1145/3366423.3380270}
\acmISBN{978-1-4503-7023-3/20/04}

\begin{document}
\title{Asking Questions the Human Way: \\Scalable Question-Answer Generation from Text Corpus} 

\author{Bang Liu$^1$, Haojie Wei$^2$, Di Niu$^1$, Haolan Chen$^2$, Yancheng He$^2$}
       \affiliation{$^1$University of Alberta, Edmonton, AB, Canada}
 \affiliation{$^2$Platform and Content Group, Tencent, Shenzhen, China}

\begin{abstract}
The ability to ask questions is important in both human and machine intelligence. Learning to ask questions helps knowledge acquisition, improves question-answering and machine reading comprehension tasks, and helps a chatbot to keep the conversation flowing with a human. Existing question generation models are ineffective at generating a large amount of high-quality question-answer pairs from unstructured text, since given an answer and an input passage, question generation is inherently a one-to-many mapping. In this paper, we propose Answer-Clue-Style-aware Question Generation (ACS-QG), which aims at automatically generating high-quality and diverse question-answer pairs from unlabeled text corpus at scale by imitating the way a human asks questions. Our system consists of: i) an information extractor, which samples from the text multiple types of assistive information to guide question generation; ii) neural question generators, which generate diverse and controllable questions, leveraging the extracted assistive information; and iii) a neural quality controller, which removes low-quality generated data based on text entailment. We compare our question generation models with existing approaches and resort to voluntary human evaluation to assess the quality of the generated question-answer pairs. The evaluation results suggest that our system dramatically outperforms state-of-the-art neural question generation models in terms of the generation quality, while being scalable in the meantime. With models trained on a relatively smaller amount of data, we can generate 2.8 million quality-assured question-answer pairs from a million sentences found in Wikipedia. 
\end{abstract}

\ccsdesc[500]{Computing methodologies~Natural language processing}
\ccsdesc[500]{Computing methodologies~Natural language generation}
\ccsdesc[100]{Computing methodologies~Machine translation}

\keywords{Question Generation, Sequence-to-Sequence, Machine Reading Comprehension}

\maketitle

\section{Introduction}
\label{sec:intro}

Automatically generating question-answer pairs from unlabeled text passages is of great value  to many applications, such as assisting the training of machine reading comprehension systems \cite{tang2017question,tang2018learning,du2018harvesting}, generating queries/questions from documents to improve search engines \cite{han2019query}, training chatbots to get and keep a conversation going \cite{shum2018eliza}, generating exercises for educational purposes \cite{heilman2010good,heilman2011automatic,danon2017syntactic}, and generating FAQs for web documents \cite{krishna2019generating}.
Many question-answering tasks such as machine reading comprehension and chatbots require a large amount of labeled samples for supervised training, acquiring which is time-consuming and costly.
Automatic question-answer generation makes it possible to provide these systems with scalable training data and to transfer a pre-trained model to new domains that lack manually labeled training samples.


\begin{figure}[tb]
\centering
\includegraphics[width=3.35in]{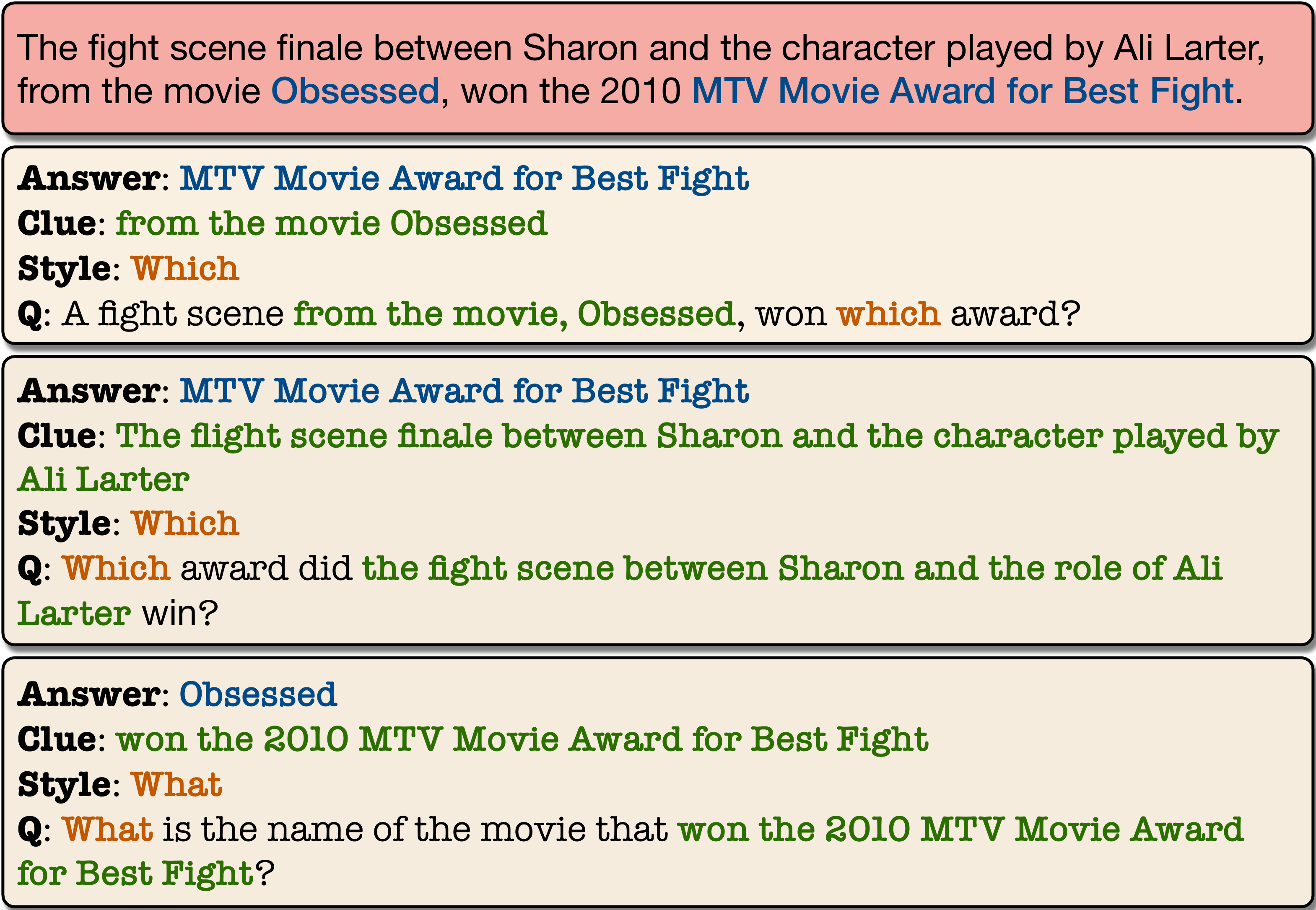}
\vspace{-5mm} 
\caption{Given the same input sentence, we can ask diverse questions based on the different choices about i) what the target answer is; ii) which answer-related chunk is used as a clue, and iii) what type of questions is asked.
}
\label{fig:FQG}
\vspace{-4mm} 
\end{figure}

Despite a large number of studies on Neural Question Generation, it remains a significant challenge to generate high-quality QA pairs from unstructured text at large quantities.
Most existing neural question generation approaches try to solve the answer-aware question generation problem, where an answer chunk and the surrounding passage are provided as an input to the model while the output is the question to be generated.
They formulate the task as a Sequence-to-Sequence (Seq2Seq) problem, and design various encoder, decoder, and input features to improve the quality of generated questions \cite{serban2016generating,du2017learning,liu2019learning,zhou2017neural,song2018leveraging,hu2018aspect,du2018harvesting}.
However, answer-aware question generation models are far from sufficient, since question generation from a passage is inherently a one-to-many mapping.
Figure~\ref{fig:FQG} shows an example of this phenomenon. Given the same input text ``The fight scene finale between Sharon and the character played by Ali Larter, from the movie Obsessed, won the 2010 MTV Movie Award for Best Fight.'', we can ask a variety of questions based on it. If we select the text chunk ``MTV Movie Award for Best Fight'' as the answer, we can still ask different questions such as ``A fight scene from the movie, Obsessed, won which award?'' or ``Which award did the fight scene between Sharon and the role of Ali Larter win?''.

We argue that when a human asks a question based on a passage, she will consider various factors.
First, she will still select an answer as a target that her question points to.
Second, she will decide which piece of information will be present (or rephrased) in her question to set constraints or context for the question. We call this piece of information as the \emph{clue}. The target answer may be related to different clues in the passage.
Third, even the same question may be expressed in different styles (e.g., ``what'', ``who'', ``why'', etc.). For example, one can ask ``which award'' or ``what is the name of the award'' to express the same meaning.
Once the answer, clue, and question style are selected, the question generation process will be narrowed down and become closer to a one-to-one mapping problem, essentially mimicking the human way of asking questions. In other words, introducing these pieces of information into question-answer generation can help reduce the difficulty of the task.

In this paper, we propose Answer-Clue-Style-aware Question Generation (ACS-QG) designed for scalable generation of high-quality question-answer pairs from unlabeled text corpus.
Just as a human will ask a question with clue and style in mind, our system first automatically extracts multiple types of information from an input passage to assist question generation. Based on the multi-aspect information extracted, we design neural network models to generate diverse questions in a controllable way.
Compared with existing answer-aware question generation, our approach essentially converts the one-to-many mapping problem into a one-to-one mapping problem, and is thus scalable by varying the assistive information fed to the neural network while in the meantime ensuring generation quality. 
Specifically, we have solved multiple challenges in the ACS-aware question generation system:


\textbf{\emph{What to ask given an unlabeled passage?}}
Given an input passage such as a sentence, randomly sampling \textit{<answer, clue, style>} combinations will cause type mismatches, since answer, clue, and style are not independent of each other. Without taking their correlations into account, for example, we may select ``how'' or ``when'' as the target question style while a person's name is selected as the answer. 
Moreover, randomly sampling \textit{<answer, clue, style>} combinations may lead to input volume explosion, as most of such combinations point to meaningless questions.

To overcome these challenges, we design and implement an information extractor to efficiently sample meaningful inputs from the given text.
We learn the joint distribution of \textit{<answer, clue, style>} tuples from existing reading comprehension datasets, such as SQuAD \cite{rajpurkar2016squad}.
In the meantime, we decompose the joint probability distribution of the tuple into three components, and apply a three-step sampling mechanism to select reasonable combinations of input information from the input passage to feed into the ACS-aware question generator. Based on this strategy, we can alleviate type mismatches and avoid meaningless combinations of  assistive information.

\textbf{\emph{How to learn a model to ask ACS-aware questions?}}
Most existing neural approaches are designed for answer-aware question generation, while there is no training data available for the ACS-aware question generation task.
We propose effective strategies to automatically construct training samples from existing reading comprehension datasets without any human labeling effort.
We define ``clue'' as a semantic chunk in an input passage that will be included (or rephrased) in the target question.
Based on this definition, we perform syntactic parsing and chunking on input text, and select the chunk which is most relevant to the target question as the clue.
Furthermore, we categorize different questions into 9 styles, including ``what'', ``how'', ``yes-no'' and so forth, 
In this manner, we have leveraged the abundance of reading comprehension datasets to automatically construct training data for ACS-aware question generation models.

We propose two deep neural network models for ACS-aware question generation, and show their superior performance in generating diverse and high-quality questions.
The first model employs sequence-to-sequence framework with copy and attention mechanism \cite{sutskever2014sequence,bahdanau2014neural,cao2017joint}, incorporating the information of answer, clue and style into the encoder and decoder. Furthermore, it discriminates between content words and function words in the input, and utilizes vocabulary reduction (which downsizes the vocabularies for both the encoder and decoder) to encourage aggressive copying.
In the second model, we fine-tune a GPT2-small model \cite{radford2019language}. We train our ACS-aware QG models using the input passage, answer, clue, and question style as the language modeling context. As a result, we reduce the phenomenon of repeating output words, which usually exists with sequence-to-sequence models, and can generate questions with better readability.
With multi-aspect assistive information, our models are able to ask a variety of high-quality questions based on an input passage, while making the generation process controllable.

\textbf{\emph{How to ensure the quality of generated QA pairs?}}
We construct a data filter, which consists of an entailment model and a question answering model.
In our filtering process, we input questions generated in the aforementioned manner into a BERT-based \cite{devlin2018bert} question answering model to get its predicted answer span, and measure the overlap between the input answer span and the predicted answer span.
In addition, we also classify the entailment relationship between the original sentence and the question-answer concatenation.
These components allow us to remove low-quality QA pairs.
By combining the input sampler, ACS-aware question generator, and the data filter, we have constructed a pipeline that is able to generate a large number of QA pairs from unlabeled text without extra human labeling efforts.

\begin{figure*}[tb]
\centering
\includegraphics[width=5.3in]{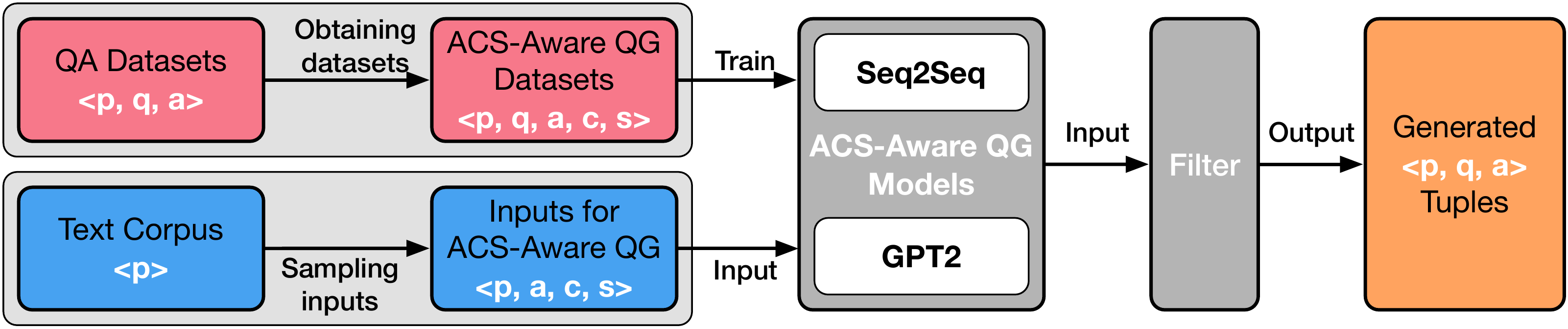}
\vspace{-2mm} 
\caption{An overview of the system architecture. It contains a dataset constructor, information sampler, ACS-aware question generator and a data filter.}
\label{fig:system}
\vspace{-4mm} 
\end{figure*}

We perform extensive experiments based on the SQuAD dataset \cite{rajpurkar2016squad} and Wikipedia, and compare our ACS-aware question generation model with different existing approaches.
Results show that both the content-separated seq2seq model with aggressive copying mechanism and the extra input information bring substantial benefits to question generation. Our method outperforms the state-of-the-art models significantly in terms of various metrics such as BLEU-4, ROUGE-L and METEOR.

With models trained on $86,635$ of SQuAD data samples, we can automatically generate two large datasets containing 1.33 million and 1.45 million QA pairs from a corpus of top-ranked  Wikipedia articles, respectively.
We perform quality evaluation on the generated datasets and identify their strengths and weaknesses.
Finally, we also evaluate how our generated QA data perform in training question-answering models in machine reading comprehension, as an alternative means to assess the data generation quality.

\section{Problem Formulation}
\label{sec:data}

In this section, we formally introduce the problem of ACS-aware question generation.

Denote a passage by $p$, where it can be either a sentence or a paragraph (in our work, it is a sentence).
Let $q$ denotes a question related to this passage, and $a$ denotes the answer of that question.
A passage consists of a sequence of words $p = \{p_t\}_{t=1}^{|p|}$ where $|p|$ denotes the length of $p$. A question $q = \{q_t\}_{t=1}^{|q|}$ contains words from either a predefined vocabulary $V$ or from the input text $p$.
Our objective is to generate different question-answer pairs from it. Therefore, we aim to model the probability $P(q, a|p)$.

In our work, we factorize the generation process into multiple steps to select different inputs for question generation. Specifically, given a passage $p$, we will select three types of information as input to a generative model, which are defined as follows:

\begin{itemize}
	\item \textbf{Answer}: here we define an answer $a$ as a span in the input passage $p$. Specifically, we select a semantic chunk of $p$ as a target answer from a set of chunks given by parsing and chunking.
	\item \textbf{Clue}: denote a clue as $c$. As mentioned in Sec.~\ref{sec:intro}, a clue is a semantic chunk in input $p$ which will be copied or rephrased in the target question. It is related to the answer $a$, and providing it as input can help reducing the uncertainty when generating questions.
	This helps to alleviate the one-to-many mapping problem of question generation, makes the generation process more controllable, as well as improves the quality of generated questions.
	\item \textbf{Style}: denote a question style as $s$. We classify each question into nine styles: \textit{``who'', ``where'', ``when'', ``why'', ``which'', ``what'', ``how'',
    ``yes-no''}, and \textit{``other''}. By providing the target style to the question generation model, we further reduce the uncertainty of generation and increase the controllability. 
\end{itemize}

We shall note that our definition of clue is different with \cite{liu2019learning}. In our work, given a passage $p$ and a question $q$, we identify a clue as a consistent chunk in $p$ instead of being the overlapping non-stop words between $p$ and $q$. On one hand, this allows the clue to be expressed in different ways in a question. On the other hand, given unlabeled text corpus, we can sample clue chunks for generating questions according to the same distribution in training datasets to avoid discrepancy between training and generating.

Given the above definitions, our generation process is decomposed into input sampling and ACS-aware question generation:

\begin{align}
\label{eq:problem}
P(q, a|p) &=  \sum_{c, s} P(a, c, s| p) P(q | a, c, s, p) \\
 &=  \sum_{c, s} P(a|p)  P(s | a, p)  P (c | s, a, p)  P (q | c, s, a, p),
\end{align}
where $P(a|p)$, $P(s | a, p)$, and $P (c | s, a, p)$ model the process of input sampling to get answer, question style, and clue information for a target question; and $P (q | c, s, a, p)$ models the process of generating the target question.

\section{Model Description}
\label{sec:model}

In this section, we present our overall system architecture for generating questions from unlabeled text corpus. We then introduce the details of each component.

Figure~\ref{fig:system} shows the pipelined system we build to train ACS-aware question generation models and generate large-scale datasets which can be utilized for different applications.
Our system consists of four major components: i) dataset constructor, which takes existing QA datasets as input, and constructs training datasets for ACS-aware question generation; ii) information sampler (extractor), which samples answer, clue and style information from input text and feed them into ACS-aware QG models; iii) ACS-aware question generation models, which are trained  on the constructed datasets to generate questions; and iv) data filter, which controls the quality of generated questions.

\subsection{Obtaining Training Data for Question Generation}
\label{subsec:obtain-data}

Our first step is to acquire a training dataset to train  ACS-aware question generation models.
Existing answer-aware question generation methods \cite{serban2016generating,du2017learning,liu2019learning,zhou2017neural} utilize reading comprehension datasets such as SQuAD \cite{rajpurkar2016squad}, as these datasets contain $<p, q, a>$ tuples.
However, for our problem, the input and output consists of $<p, q, a, c, s>$, where the clue $c$ and style $s$ information are not directly provided in existing datasets.
To address this issue, we design effective strategies to automatically extract clue and style information without involving human labeling.

\begin{algorithm}[h]
\KwIn{passage $p$, answer $a$, question $q$, related words dictionary $R$.}
\KwOut{ clue $c$.}
\begin{algorithmic}[1]
\STATE get candidate chunks $C = \{c_1, c_2, \cdots, c_{|C|}\}$ of passage $p$ by parsing and chunking\;
\STATE remove function words, tokenize $p$ and $q$ to get $p_{t,c}$ and $q_{t,c}$\, and stemming $p$ and $q$ to get $p_{m,c}$ and $q_{m,c}$\;

\FOR{$c \in C$}
\STATE get tokenized clue $c_{t,c}$ and stemmed clue $c_{m,c}$ with only content words\;
\STATE $n^o_{t,c} \leftarrow$ number of overlapping  tokens between $c_{t,c}$ and $q_{t,c}$\;
\STATE $n^o_{m,c} \leftarrow$ number of overlapping stems between $c_{m,c}$ and $q_{m,c}$\;
\STATE $n^{soft-o}_{t,c} \leftarrow$ number of soft copied tokens between $c_{t,c}$ and $q_{t,c}$\;
\STATE binary $x \leftarrow$ whether $q$ contains the chunk text $c$\;
\STATE $score(c) = n^o_{t,c} + n^o_{m,c} + n^{soft-o}_{t,c} + x$
\ENDFOR
\STATE select the chunk $c$ with maximum $score(c)$ as the clue chunk\;
\end{algorithmic}
\caption{Clue Extraction}
\label{alg:clue-extract}
\end{algorithm}

\textbf{Rules for Clue Identification}.
As mentioned in Sec.~\ref{sec:data}, given $<p, q, a>$, we define a semantic chunk $c$ in input $p$ which is copied or rephrased in the output question $q$ as clue.
We identify $c$ by the method shown in Algorithm~\ref{alg:clue-extract}.

First, we parse and chunk the input passage to get all candidate chunks. Second, we get the tokenized and stemmed passage and question, and only keep the content words in the results. Third, we calculate the similarities between each candidate chunk and the target question according to different criteria. The final score of each chunk is the sum of different similarities. Finally, we select the chunk with the maximum score as the identified clue chunk $c$.

To estimate the similarities between each candidate chunk and the question, we calculate the number of overlapping tokens and stems between each chunk and the question, as well as checking whether the chunk is fully contained in the question. In addition, we further define ``soft copy'' relationship between two words to take rephrasing into consideration. Specifically, a word $w_q \in q$ is considered as \textit{soft-copied} from input passage $p$ if there exist a word $w_p \in p$ which is semantically coherent with $w_q$. To give an instance, consider a passage ``Selina left her hometown at the age of 18'' and a question ``How old was Selina when she left?'', the word ``old'' is soft-copied from ``age'' in the input passage.

To identify the soft-copy relationship between any pair of words, we utilize synonyms and word vectors, such as Glove \cite{pennington2014glove}, to construct a related words dictionary $R$, where $R(w) = \{w_1, w_2, \cdots, w_{|R(w)|}\}$ returns a set of words that is closely related to $w$. For each word $w$, $R(w)$ is composed of the synonyms of $w$, as well as the top $N$ most similar words estimated by word vector representations (we set $N = 5$). In our work, we utilize Glove word vectors, and construct $R$ based on Genism \cite{rehurek_lrec} and WordNet \cite{miller1995wordnet}.

\textbf{Rules for Style Classification}.
Algorithm~\ref{alg:style-extract} presents our method for question style classification. We classify a given question into $9$ classes based on a few heuristic strategies. If $q$ contains \textit{who, where, when, why, which, what}, or \textit{how}, we classify it as the corresponding type. For \textit{yes-no} type questions, we define a set of feature words. If $q$ starts with any word belonging to the set of feature words, we classify it as type \textit{yes-no}. For all other cases, we label it as \textit{other}.

\begin{algorithm}[h]
\KwIn{question $q$, style set $S = \{who, where, when, why, which, what, how, yes\text{-}no, other\}$, yes-no feature words set $Y = \{am, is, was, were, are, does, do, did, have, had, has, could, \newline can, shall, should, will, would, may, might\}.$}
\KwOut{ style $s \in S$.}

\begin{algorithmic}[1]
\FOR{$s \in S\backslash \{yes\text{-}no, other\}$}
\IF {word $s$ is contained in $q$}
    \RETURN{$s$} 
\ENDIF
\ENDFOR
\FOR{$y \in Y$}
\IF {word $y$ is the first word of $q$}
    \RETURN{$yes\text{-}no$} 
\ENDIF
\ENDFOR
\RETURN{$other$}
\end{algorithmic}
\caption{Style Classification}
\label{alg:style-extract}
\end{algorithm}


\subsection{ACS-Aware Question Generation}
\label{subsec:train-model}


After obtained training datasets, we design two models for ACS-aware question generation. The first model is based on Seq2Seq framework with attention and copy mechanism \cite{sutskever2014sequence,bahdanau2014neural,gu2016incorporating}. In addition, we exploit clue embedding, content embedding, style encoding and aggressive copying to improve the performance of question generation. The second model is based on pre-trained language models. We fine-tune a GPT2-small model \cite{radford2019language} using the constructed training datasets.

\subsubsection{Seq2Seq-based ACS-aware question generation}
\label{subsubsec:seq2seq}

Given a passage, an answer span, a clue span, and a desired question style, we train a neural encoder-decoder model to generate appropriate questions.

\textbf{Encoder}. We utilize a bidirectional Gated Recurrent Unit (BiGRU) \cite{chung2014empirical} as our encoder. For each word $p_i$ in input passage $p$, we concatenate the different features to form a concatenated embedding vector $w_i$ as the input to the encoder.
Specifically, for each word, it is represented by the concatenation of its word vector, embeddings of its Named Entity Recognition (NER) tag, Part-of-Speech (POS) tag, and whether it is a content word. In addition, we can know whether each word is within the span of answer $a$ or clue $c$, and utilize binary features to indicate the positions of answer and clue in input passage. All tag features and binary features are casted into $16$-dimensional vectors by different embedding matrices that are trainable.

Suppose the embedding of passage $p$ is $(w_1, w_2, \cdots, w_{|p|})$. Our encoder will read the input sequence and produce a sequence of hidden states $h_1, h_2, \cdots, h_{|p|}$, where each hidden state is a concatenation of a forward representation and a backward representation:
\begin{align}
\label{eq:bigru}
h_i &= [\overrightarrow{h}_i; \overleftarrow{h}_i],\\
\overrightarrow{h}_i &= \mathbf{BiGRU}(w_i, \overrightarrow{h}_{i-1}),\\
\overleftarrow{h}_i &= \mathbf{BiGRU}(w_i, \overleftarrow{h}_{i+1}).
\end{align}
The $\overrightarrow{h}_i$ and $\overleftarrow{h}_i$ are the forward and backward hidden states of the $i$-th token in $p$, respectively.

\textbf{Decoder}. Our decoder is another GRU with attention and copy mechanism. Denote the embedding vector of desired question style $s$ as $h_s$. We initialize the hidden state of our decoder GRU by concatenating $h_s$ with the last backward encoder hidden state $\overleftarrow{h}_1$ to a linear layer:
\begin{align}
\label{eq:decinit}
s_l &= tanh(\textbf{W}_0\overleftarrow{h}_1 + b),\\
s_0 &= [h_s; s_l].
\end{align}

At each decoding time step $t$, the decoder calculates its current hidden state based on the word vector of the previous predicted word $w_{t-1}$, previous attentional context vector $c_{t-1}$, and its previous hidden state $s_{t-1}$:
\begin{align}
\label{eq:decoder-update}
s_t = \mathbf{GRU}([w_{t-1}; c_{t-1}], s_{t-1}),
\end{align}
where the context vector $c_t$ at time step $t$ is a weighted sum of input hidden states, and the weights are calculated by the concatenated attention mechanism \cite{luong2015effective}:
\begin{align}
\label{eq:attention}
e_{t,i} &= v^\intercal tanh(\mathbf{W}_s s_{t} + \mathbf{W}_h h_i),\\
\alpha_{t,i} &= \frac{\text{exp}(e_{t,i})}{\sum_{j=1}^{|p|}\text{exp}(e_{t,j})},\\
c_t &= \sum_{i=1}^{|p|}\alpha_{t,i} h_i.
\end{align}

To generate an output word, we combine $w_{t-1}$, $s_t$ and $c_t$ to calculate a readout state $r_t$ by an MLP maxout layer with dropouts \cite{goodfellow2013maxout}, and pass it to a linear layer and a softmax layer to predict the probabilities of the next word over a vocabulary:
\begin{align}
\label{eq:generator}
r_t &= \mathbf{W}_{rw} w_{t-1} + \mathbf{W}_{rc} c_t + \mathbf{W}_{rs} s_t\\
m_t &= [\text{max}\{r_{t,2j-1}, r_{t, 2j}\}]_{j=1,...,d}^\intercal\\
p(y_t|& y1,\cdots, y_{t-1}) = \text{softmax}(\mathbf{W}_o m_t),
\end{align}
where $r_t$ is a $2$-D vector.

For copy or point mechanism \cite{gulcehre2016pointing},
the probability to copy a word from input $p$ at time step $t$ is given by:
\begin{align}
g_c = \sigma(\mathbf{W}_{cs} s_t + \mathbf{W}_{cc} c_t + b),
\end{align}
where $\sigma$ is the Sigmoid function, and $g_c$ is the probability of performing copying.
The copy probability of each input word is given by the attention weights in Equation (10).

It has been reported that the generated words in a target question are usually from frequent words, while the majority of low-frequency words in the long tail are copied from the input instead of generated  \cite{liu2019learning}. Therefore, we reduce the vocabulary size to be the top $N_V$ high-frequency words at both the encoder and the decoder, where $N_V$ is a predefined threshold that varies among different datasets. This  helps to  encourage the model to learn aggressive copying and improves the performance of question generation.

\subsubsection{GPT2-based ACS-aware question generation}
\label{subsubsec:gpt2}

\begin{figure}[tb]
\centering
\includegraphics[width=3.2in]{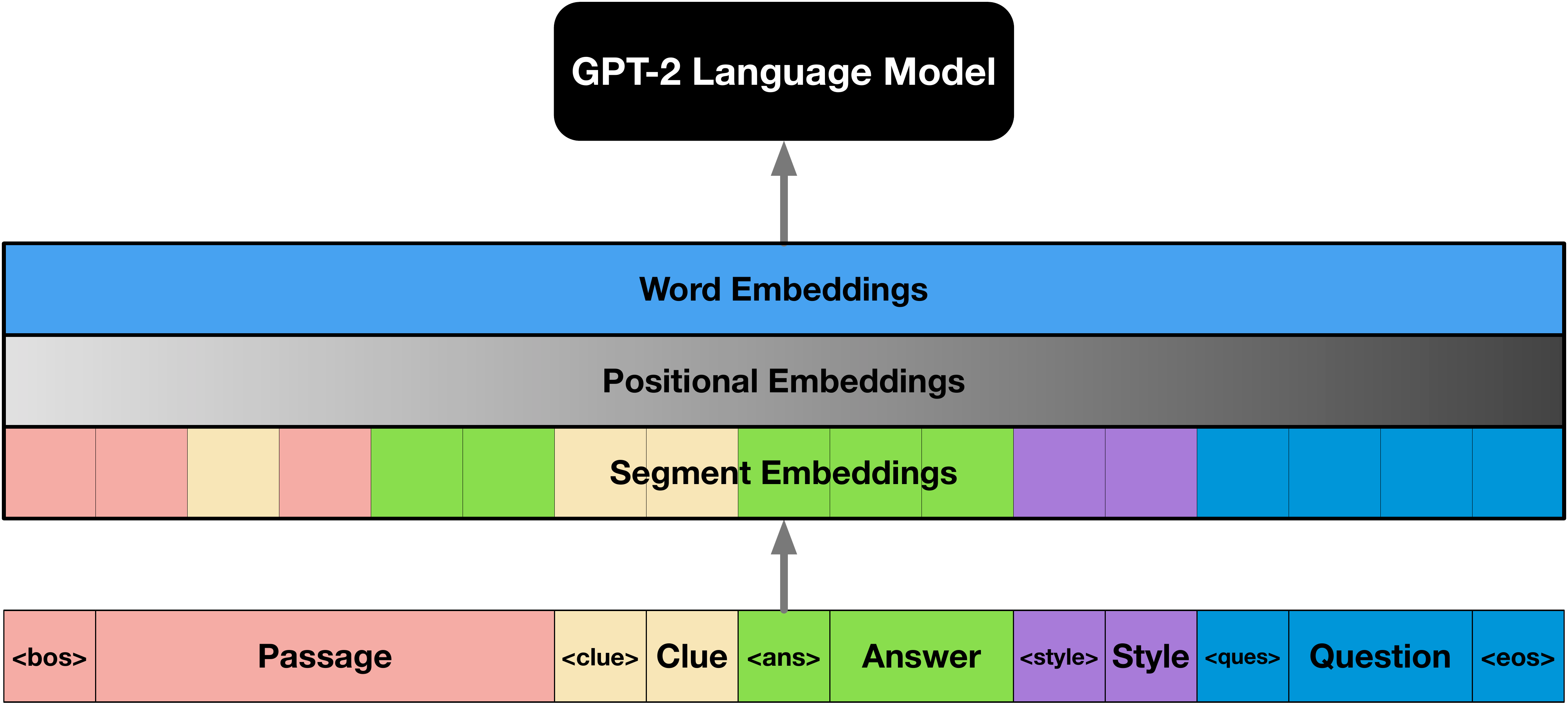}
\vspace{-2mm} 
\caption{The input representations we utilized for fine-tuning GPT-2 Transformer-based language model.
}
\label{fig:gpt2}
\vspace{-4mm} 
\end{figure}

Pre-trained large-scale language models, such as BERT \cite{devlin2018bert}, GPT2 \cite{radford2019language} and XLNet \cite{yang2019xlnet}, have significantly boosted the performance of a series of NLP tasks including text generation.
They are mostly based on the Transformer architecture \cite{vaswani2017attention} and have been shown to capture many facets of language relevant for downstream tasks \cite{clark2019does}.
Compared with Seq2Seq models which often generate text containing repeated words, the pre-trained language models acquire knowledge from large-scale training dataset and are able to generate text of high-quality.
In our work, to produce questions with better quality and to compare with Seq2Seq-based models, we further fine-tune the publicly-available pre-trained GPT2-small model \cite{radford2019language} according to our problem settings.

Specifically, to obtain an ACS-question generation model with GPT2, we concatenate passage, answer, clue and question style as the context of language modeling.
Specifically, the input sequence is organized in the form of ``\textit{<bos> ..passage text.. <clue> .. clue chunk .. <ans> .. answer chunk .. <style> .. question style .. <ques> .. question text .. <eos>}''.
During the training process, we learn a language model with above input format.
When generating, we sample different output questions by starting with an input in the form of ``\textit{<bos> ..passage text.. <clue> .. clue chunk .. <ans> .. answer chunk .. <style> .. question style .. <ques>}''.
Figure~\ref{fig:gpt2} illustrates the input representation for fine-tuning GPT-2 language model.
Similar to \cite{krishna2019generating}, we leverage GPT-2's segment embeddings to denote the specificity of the passage, clue, answer, style and question.
We also utilize answer segment embeddings and clue segment embedding in place of passage segment embeddings at the location of the answer or clue in the passage to denote the position of the answer span and clue span.
During the generation process, the trained model uses top-p nucleus sampling with $p = 0.9$ \cite{holtzman2019curious} instead of beam search and top-k sampling.
For implementation, we utilize the  code base of \cite{krishna2019generating} as a starting point, as well as the Transformers library from HuggingFace \cite{wolf2019transformers}.

\subsection{Sampling Inputs for Question Generation}
\label{subsec:sample-data}

As mentioned in Sec.~\ref{sec:data}, the process of ACS-aware question generation consists of input sampling and text generation.
Given an unlabeled text corpus, we need to extract valid \textit{<passage, answer, clue, style>} combinations as inputs to generate questions with an ACS-aware question generation model.

In our work, we decompose the sampling process into three steps to sequentially sample the candidate answer, style and clue based on a given passage.
We make the following assumptions: i) the probability of a chunk $a$ be selected as an answer only depends on its Part-of-Speech (POS) tag, Named Entity Recognition (NER) tag and the length (number of words) of the chunk; ii) the style $s$ of the target question only depends on the POS tag and NER tag of the selected answer $a$; and iii) the probability of selecting a chunk $c$ as clue depends on the POS tag and the NER tag of $c$, as well as the dependency distance between chunk $c$ and $a$.
We calculate the length of the shortest path between the first word of $c$ and that of $a$ as the dependency distance.
The intuition for the last designation is that a clue chunk is usually closely related to the answer, and is often copied or rephrased into the target question. Therefore, the dependency distance between a clue and an answer will not be large \cite{liu2019learning}.

With above assumptions, we will have:

\begin{align}
\label{eq:problem}
P(a | p) &=  P(a| POS(a), NER(a), length(a)), \\
P(s | a, p) &= P(s | POS(a), NER(a)), \\
P(c | s, a, p) &= P(c | POS(c), NER(c), DepDist(c, a)),
\end{align}
where $DepDist(c, a)$ represents the dependency distance between the first token of $c$ and that of $a$.

\begin{figure}[tb]
\centering
\subfigure[Joint Distribution of Answer NER and Answer Length]{
\includegraphics[width=3.3in]{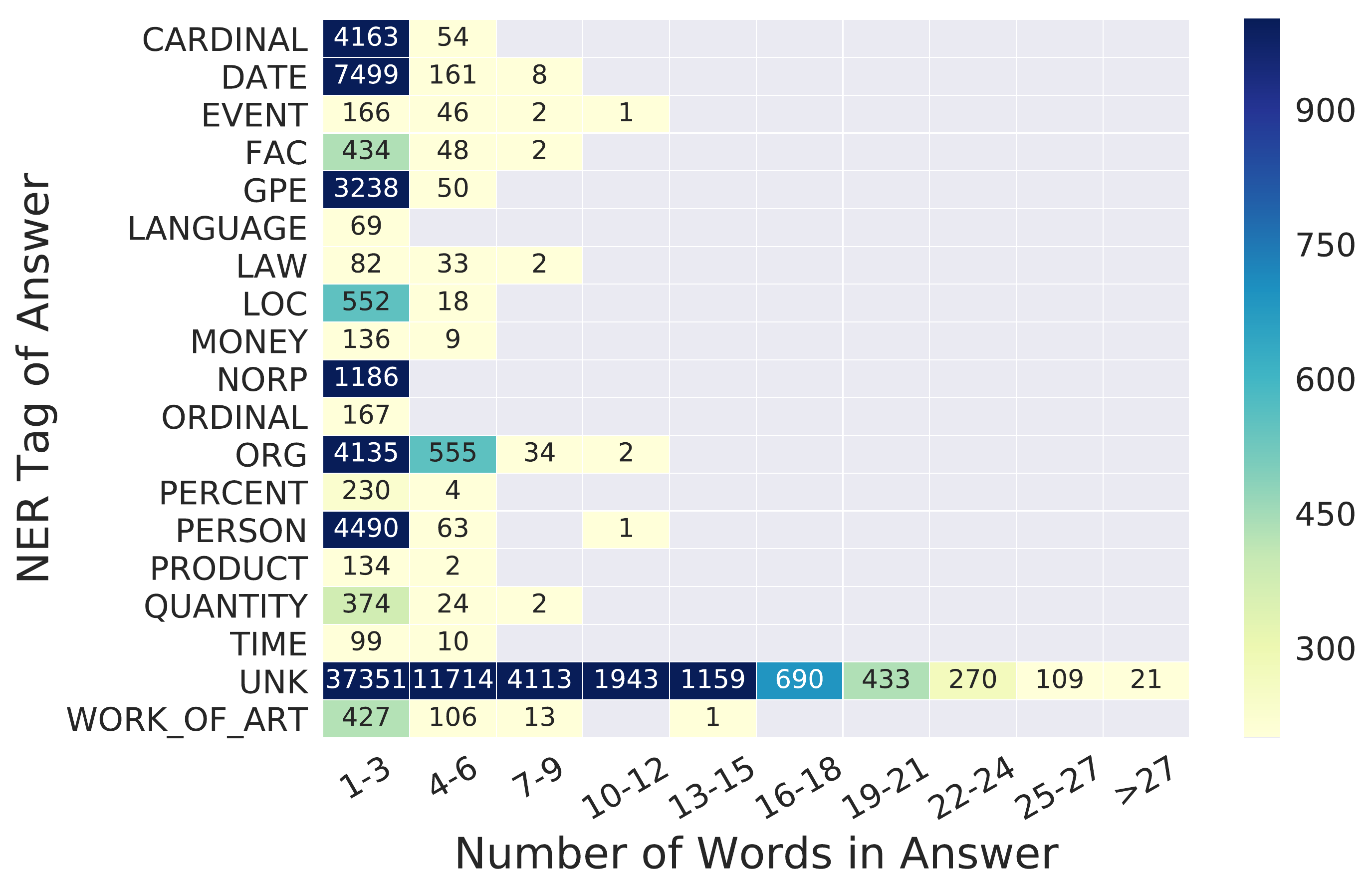}
\label{fig:answer}
}
\vspace{-1mm} 
\subfigure[Joint Distribution of Clue NER and Dependency Distance]{
\includegraphics[width=3.3in]{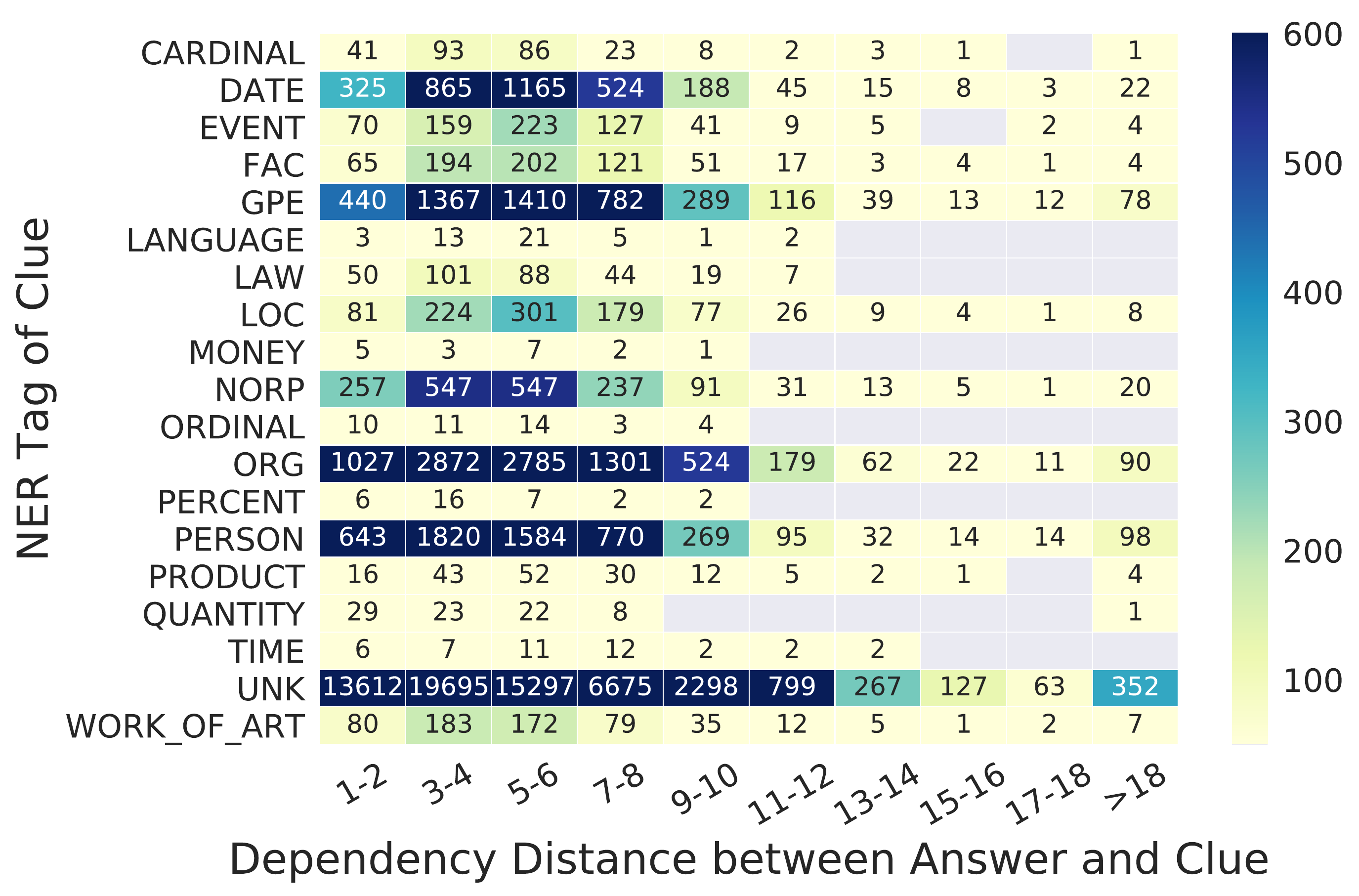}
\label{fig:clue}
}
\vspace{-1mm} 
\subfigure[Joint Distribution of Answer NER and Question Style]{
\includegraphics[width=3.3in]{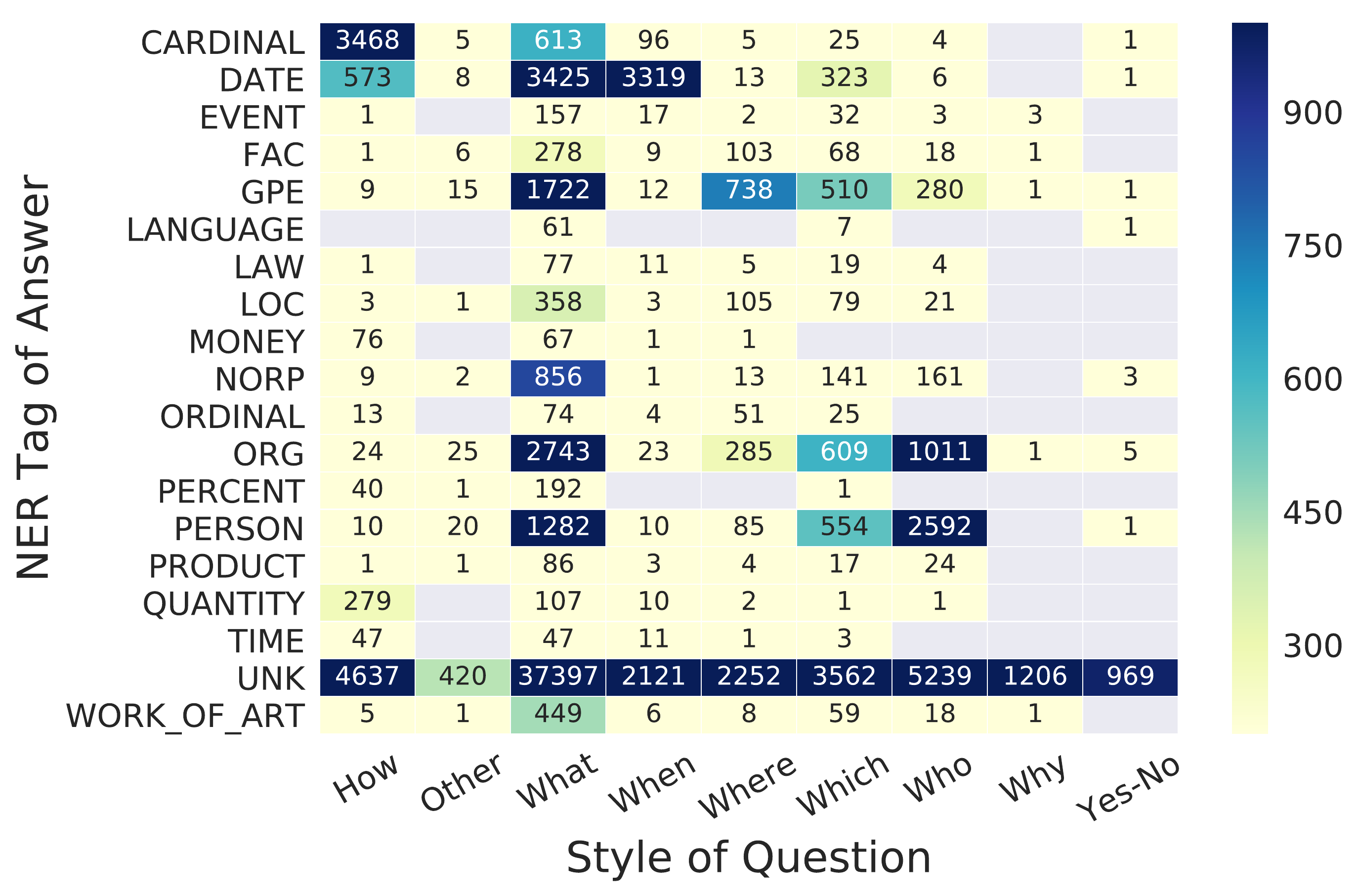}
\label{fig:style}
}
\vspace{-3mm} 
\caption{The input join distributions we get using SQuAD1.1 training dataset as reference data.}
\label{fig:squad-info}
\vspace{-5mm} 
\end{figure}

The above conditional probabilistic distributions can be learned from an existing dataset (such as SQuAD), named as reference dataset.
Given a reference dataset consisting of \textit{<passage, question, answer>} triplets, first, we perform POS tagging, NER tagging, parsing and chunking on the input passages.
Second, we recognize the clue and style information according to the steps described in Sec.~\ref{subsec:obtain-data}, get the NER tag and the POS tag of both the answer chunk and the clue chunk, and calculate the dependency distance between the clue chunk and the answer chunk.
Finally, we calculate the conditional distributions according to the extracted information.

We set the maximum length of an candidate answer to 30, and split the range of length into 10 bins of equal size to calculate $P(a| POS(a), NER(a), length(a))$.
Similarly, we set the maximum dependency distance between a clue chunk and an answer chunk to be 20, and split the range of distance into 10 bins of equal size to calculate $P(c | POS(c), NER(c), DepDist(c, a))$.
Figure~\ref{fig:squad-info} shows the marginal distributions we get by utilizing the SQuAD1.1 training dataset as our reference data. The NER tagging \footnote{The meaning of NER labels can be found at: https://spacy.io/api/annotation} is performed by spaCy \cite{spacy2}, and the ``UNK'' tag means the chunk is not recognized as a named entity. From Figure~\ref{fig:answer}, we can see that most of the answers are short, and a majority of them are entities such as person (PERSON), organization (ORG) or date (DATE). From Figure~\ref{fig:clue}, we can see the syntactical dependency distance between a clue chunk and an answer is usually less than $8$, which matches with our intuition that a clue shall be correlated with the answer so that it will be copied or rephrased in the target question. Finally, Figure~\ref{fig:style} shows most of the questions are  ``What'' style, and the followings are ``Who'', ``How'' and ``When''. The NER tags of answers are highly correlated with the style of questions. Also, we shall notice that the NER performance of spaCy is not perfect. Therefore, we may observe weird cases such as organization (ORG) matches with ``Who''. Determining different conditional probabilities with a reference dataset instead of following pre-defined rules helps us to take such kind of noises into account.

After calculating the above distributions according to an reference dataset, we can sample different information given a passage $p$. First, we get all candidate chunks $K = \{k_1, k_2, \cdots, k_{|K|}\}$ by parsing and chunking over $p$. Second, we sample a chunk $k_i$ as answer according to the normalized probability distribution over all chunks:
\begin{align}
P(k_i) = \frac{P(k_i|POS(k_i), NER(k_i), length(k_i))}{\sum_{j=1}^{|K|} P(k_j|POS(k_j), NER(k_j), length(k_j))}.
\end{align}
Third, we sample a question style $s_i$ over all possible questions styles $S = \{s_1, s_2, \cdots, s_{|S|}\}$ by the normalized probability:
\begin{align}
P(s_i) = \frac{P(s_i|POS(k_i), NER(k_i))}{\sum_{j=1}^{|S|} P(s_j|POS(k_i), NER(k_i))}.
\end{align}
Finally, we sample a chunk $k_l$ as clue according to:
\begin{align}
P(k_l) = \frac{P(k_l|POS(k_l), NER(k_l), DepDist(k_l, k_i))}{\sum_{j=1}^{|K|} P(k_j|POS(k_j), NER(k_j), DepDist(k_j, k_i))}.
\end{align}
We can repeat the above steps for multiple times to get different inputs from the same passage and generate diverse questions. In our work, for each input passage, we sample $5$ different chunks as answer spans, $2$ different question styles for each answer, and $2$ different clues for each answer. In this way, we over-generate questions by sampling $20$ questions for each sentence.

\subsection{Data Filtering for Quality Control}
\label{subsec:filter-data}

After sampled multiple inputs from each sentence, we can generate different questions based on the inputs. However, it is hard to ask each sentence 20 meaningful and different questions even given 20 different inputs derived from it, as the questions may be duplicated due to similar inputs, or the questions can be meaningless if the \textit{<answer, clue, style>} combination is not reasonable. Therefore, we further utilize a filter to remove low-quality QA pairs.

We leverage an entailment model and a QA model based on BERT \cite{devlin2018bert}. For the entailment model, as the SQuAD 2.0 dataset \cite{rajpurkar2018know} contains unanswerable questions, we utilize it to train a classifier which tells us whether a pair of \textit{<question, answer>} matches with the content in the input passage.
For the question answering model, we fine-tuned another BERT-based QA model utilizing the SQuAD 1.1 dataset \cite{rajpurkar2016squad}.

Given a sample \textit{<passage, question, answer>}, we keep it if this sample satisfies two criteria: first, it is classified as positive according to the BERT-based entailment model; second, the F1 similarity score between the gold answer span and the answer span predicted by BERT-based QA is above 0.9. 
Note that we do not choose to fine-tune a BERT-based QA model over SQuAD 2.0 to perform entailment and question answering at the same time.
That is because we get better performance by separating the entailment step with the QA filtering step. Besides, we can utilize extra datasets from other entailment tasks to enhance the entailment model and further improve the data filter.

\section{Evaluation}
\label{sec:simu}

In this section, we compare our proposed ACS-aware question generation with answer-aware question generation models to show its benefits.
We generate a large number of QA pairs from unlabeled text corpus using our models, and further perform a variety of evaluations to analyze the quality of the generated data.
Finally, we test the performance of QA models trained on our generated dataset, and show the potential applications and future directions of this work.

\subsection{Evaluating ACS-aware Question Generation}

\textbf{Datasets, Metrics and Baselines}.
We evaluate the performance of ACS-aware question generation based on the SQuAD dataset \cite{rajpurkar2016squad}. It is a reading comprehension dataset which contains questions derived from Wikipedia articles, and the answer to every question is a segment of text from the corresponding reading passage.
In our work, we use the data split proposed by \cite{zhou2017neural}, where the input is the sentence that contains the answer. The training set contains $86,635$ samples, and the original dev set that contains $17,929$ samples is randomly split into a dev test and a test set of equal size. The average lengths (number of words) of sentences, questions and answers are $32.72$, $11.31$, and $3.19$, respectively.

The performance of question generation is evaluated by the following metrics.
\begin{itemize}
	\item \textbf{BLEU} \cite{papineni2002bleu}. 
	BLEU measures precision by how much the words in predictions appear in reference sentences. BLEU-1 (B1), BLEU-2 (B2), BLEU-3 (B3), and BLEU-4 (B4), use 1-gram to 4-gram for calculation, respectively.
	\item \textbf{ROUGE-L} \cite{lin2004rouge}.
	ROUGE-L measures recall by how much the words in reference sentences appear in predictions using Longest Common Subsequence (LCS) based statistics.
	\item \textbf{METEOR} \cite{denkowski2014meteor}. 
	METEOR is based on the harmonic mean of unigram precision and recall, with recall weighted higher than precision.
\end{itemize}

We compare our methods with the following baselines.

\begin{itemize}
	\item \textbf{PCFG-Trans} \cite{heilman2011automatic}: a rule-based answer-aware question generation system.
	\item \textbf{SeqCopyNet} \cite{zhou2018sequential}, \textbf{NQG++} \cite{zhou2017neural}, \textbf{AFPA} \cite{sun2018answer}, \textbf{seq2seq+z+c+GAN} \cite{yao2018teaching}, and \textbf{s2sa-at-mp-gsa} \cite{zhao2018paragraph}: answer-aware neural question generation models based on Seq2Seq framework.
	\item \textbf{NQG-Knowledge} \cite{gupta2019improving}, \textbf{DLPH} \cite{gao2018difficulty}: auxiliary-information-enhanced question generation models with extra inputs such as knowledge or difficulty.
	\item \textbf{Self-training-EE} \cite{sachan2018self}, \textbf{BERT-QG-QAP} \cite{zhang2019addressing}, \textbf{NQG-LM} \cite{zhou2019multi}, \textbf{CGC-QG} \cite{liu2019learning} and \textbf{QType-Predict} \cite{zhou2019type}: multi-task question generation models with auxiliary tasks such as question answering, language modeling, question type prediction and so on.
\end{itemize}

The reported performance of baselines are directly copied from their papers or evaluated by their published code on GitHub.

For our models, we evaluate the following versions:
\begin{itemize}
    \item \textbf{CS2S-VR-A}. Content separated Seq2Seq model with Vocabulary Reduction for Answer-aware question generation. In this variant, we incorporate content embeddings in word representations to indicate whether each word is a content word or a function word. Besides, we reduce the size of vocabulary by only keeping the top $2000$ frequent words for encoder and decoder. In this way, low-frequency words are represented by its NER, POS embeddings and feature embeddings. We also add answer position embedding to indicate the answer span in input passages.
    \item \textbf{CS2S-AS}. This model adds question style embedding to initialize decoder, without vocabulary reduction (vocabulary size is $20,000$ when we do not exploit vocabulary reduction).
    \item \textbf{CS2S-AC}. The variant adds clue embedding in encoder to indicate the span of clue chunk.
    \item \textbf{CS2S-ACS}. This variant adds both clue embedding in encoder and style embedding in decoder.
    \item \textbf{CS2S-VR-ACS}. This is the fully featured model with answer, clue and style embedding, as well as vocabulary reduction.
    \item \textbf{GPT2-ACS}. This is our fine-tuned GPT2-small model for ACS-aware question generation. 
\end{itemize}

\begin{table}[tb]
\tabcolsep=0.11cm
\small
  \begin{tabular}{c|cccccc}
    \toprule
    \textbf{Model} & \textbf{B1} & \textbf{B2} & \textbf{B3} & \textbf{B4} & \textbf{ROUGE-L} & \textbf{METEOR} \\
    \midrule
    PCFG-Trans & $28.77$ & $17.81$ & $12.64$ & $9.47$ & $31.68$ & $18.97$ \\
    SeqCopyNet & $-$ & $-$ & $-$ & $13.02$ & $44.00$ & $-$ \\
    seq2seq+z+c+GAN & $44.42$ & $26.03$ & $17.60$ & $13.36$ & $40.42$ & $17.70$ \\
    NQG++ & $42.36$ & $26.33$ & $18.46$ & $13.51$ & $41.60$ & $18.18$ \\
    AFPA & $43.02$ & $28.14$ & $20.51$ & $15.64$ & $-$ & $-$ \\
    s2sa-at-mp-gsa & $44.51$ & $29.07$ & $21.06$ & $15.82$ & $44.24$ & $19.67$ \\
    \hdashline
    NQG-Knowledge & $-$ & $-$ & $-$ & $13.69$ & $42.13$ & $18.50$ \\
    DLPH & $44.11$ & $29.64$ & $21.89$ & $16.68$ & $46.22$ & $20.94$ \\
    \hdashline
    NQG-LM & $42.80$ & $28.43$ & $21.08$ & $16.23$ & $-$ & $-$ \\
    QType-Predict & $43.11$ & $29.13$ & $21.39$ & $16.31$ & $-$ & $-$ \\
    Self-training-EE & $-$ & $-$ & $-$ & $14.28$ & $42.97$ & $18.79$ \\ 
    CGC-QG & $46.58$ & $30.90$ & $22.82$ & $17.55$ & $44.53$ & $21.24$ \\
    BERT-QG-QAP & $-$ & $-$ & $-$ & $18.65$ & $46.76$ & $22.91$ \\   
    \midrule
    CS2S-VR-A & $45.28$ & $29.58$ & $21.45$ & $16.13$ & $43.98$ & $20.59$\\
    CS2S-AS & $45.79$ & $29.12$ & $20.59$ & $15.09$ & $45.84$ & $20.14$\\
    CS2S-AC & $48.13$ & $32.51$ & $24.08$ & $18.40$ & $47.45$ & $22.27$\\
    CS2S-ACS & $50.72$ & $34.60$ & $25.79$ & $19.84$ & $51.08$ & $23.58$\\
    CS2S-VR-ACS & $\mathbf{52.30}$ & $\mathbf{36.70}$ & $\mathbf{28.00}$ & $\mathbf{22.05}$ & $\mathbf{53.25}$ & $25.11$\\
    GPT2-ACS & $42.60$ & $31.23$ & $24.00$ & $18.87$ & $43.63$ & $\mathbf{25.15}$\\
    \bottomrule
  \end{tabular}
  \caption{Evaluation results of different models on SQuAD dataset.}
\vspace{-7mm}
  \label{tab:squad-dataset}
\end{table}

\begin{table}[t]
\centering
\tabcolsep=0.11cm
\small
  \begin{tabular}{p{1.5cm}|c|ccc}
    \toprule
    \multicolumn{2}{c|}{\textbf{Experiments}} & \textbf{CS2S-VR-ACS} & \textbf{GPT2-ACS} & \textbf{GOLD} \\
    \hline
    \multirow{3}{*}{\parbox[t]{1.5cm}{Question is \\Well-formed}} & No & $28.5\%$ & $6.0\%$ & $2.0\%$ \\ 
    & Understandable & $31.5\%$ & $19.5\%$ & $9.0\%$  \\
    & Yes & $40.0\%$ & $74.5\%$ & $89.0\%$ \\
    \hline
    \multirow{2}{*}{\parbox[t]{1.5cm}{Question is \\ Relevant}} & No & $6.3\%$ & $11.7\%$ & $7.1\%$ \\ 
    & Yes & $93.7\%$ & $88.3\%$ & $92.9\%$  \\
    \hline
    \multirow{3}{*}{\parbox[t]{1.5cm}{Answer is \\ Correct}} & No & $7.4\%$ & $3.6\%$ & $2.2\%$ \\ 
    & Partially & $12.7\%$ & $15.1\%$ & $15.4\%$  \\
    & Yes & $79.9\%$ & $81.3\%$ & $82.4\%$  \\
    \bottomrule
  \end{tabular}
  \caption{Human evaluation results about the quality of generated QA pairs.}
  \label{tab:quality}
  \vspace{-7mm}
\end{table}

\textbf{Experiment Settings}.
We implement our models in PyTorch 1.1.0 \cite{paszke2017pytorch} and Transformers 2.0.0 \cite{wolf2019transformers}, and train the model with two Tesla P40. We utilize spaCy \cite{spacy} to perform dependency parsing and extract lexical features for tokens.
For Seq2Seq-based models, we set word embeddings to be $300$-dimensional and initialize them by GloVe, and set them trainable.
The out-of-vocabulary words are initialized randomly. 
All other features are embedded to 16-dimensional vectors.
The encoder is a single layer BiGRU with hidden size 512, and the decoder is a single layer undirected GRU with hidden size 512. 
We set dropout rate $p = 0.1$ for the encoder, decoder, and the attention module.
We train the models by Cross-Entropy loss for question generation and question copying, and perform gradient descent by the Adam \cite{kingma2014adam} optimizer with an initial learning rate $l_r =0.001$, two momentum parameters are $\beta_1=0.8$ and $\beta_2=0.999$ respectively, and $\epsilon=10^{-8}$. The mini-batch size for each update is set to 32 and model is trained for up to 10 epochs. Gradient clipping with range $[-5, 5]$ is applied to Adam.
Beam width is set to be $20$ for decoding. The decoding process stops when the \textit{<EOS>} token (represents end-of-sentence) is generated.

For GPT2-ACS model, we fine-tune the GPT2-small model using SQuAD 1.1 training dataset from \cite{zhou2017neural}. We fine-tune the model for 4 epochs with batch size 2, and apply top-$p$ nucleus sampling with $p = 0.9$ when decoding. For BERT-based filter, we fine-tune the BERT-large-uncased  model from HuggingFace \cite{wolf2019transformers} with parameters suggested by \cite{wolf2019transformers} for training on SQuAD 1.1 and SQuAD 2.0.
Our code will be published for research purpose\footnote{https://github.com/bangliu/ACS-QG}.

\textbf{Main Results}.
Table.~\ref{tab:squad-dataset} compares our models with  baseline approaches.
We can see that our CS2S-VR-ACS achieves the best performance in terms of the evaluation metrics and outperforms baselines by a great margin.
Comparing CS2S-VR-A with Seq2Seq-based answer-aware QG baselines, we can see that it outperforms all the baseline approaches in that category with the same input information (input passage and answer span).
This is because that our content separation strategy and vocabulary reduction operation help the model to better learn what words to copy from the inputs.
Comparing our ACS-aware QG models and variants with auxiliary-information-enhanced models (such as NQG-Knowledge and DLPH) and auxiliary-task-enhanced baselines (such as BERT-QG-QAP), we can see that the clue and style information helps to generate better results than models with knowledge or difficulty information.
That is because our ACS-aware setting makes the question generation problem closer to one-to-one mapping, and greatly reduces the task difficulty.

Comparing GPT2-ACS with CS2S-VR-ACS, we can see that GPT2-ACS achieves better METEOR score, while CS2S-VR-ACS performs better over BLEU scores and ROUGE-L. That is because GPT2-ACS has no vocabulary reduction. Hence, the generated words are more flexible. However, metrics such as BLEU scores are not able to evaluate the quality of generated QA pairs semantically. Therefore, in the following section, we further analyze the quality of  QA pairs generated by CS2S-VR-ACS and GPT2-ACS to identify their strengths and weaknesses.

\subsection{Qualitative Analysis}

After training our CS2S-VR-ACS and GPT2-ACS models, we generate large-scale \textit{<passage, question, answer>} datasets from unlabeled text corpus.
Specifically, we obtain the top $10,000$ English Wikipedia articles with Project Nayuki's Wikipedia's internal PageRanks. After that, we split each article in the corpus into sentences, and filter out sentences with lengths shorter than $5$ or longer than $100$.
Based on these sentences, we perform input sampling to sample \textit{<answer, clue, style>} triplets for each sentence according to the steps described in Section~\ref{subsec:sample-data}, and feed them into our models to generate questions. After filtering, we create two datasets utilizing the two models, where each of the dataset contains around $1.4$ million generated questions.

We first evaluate the quality of the generated QA pairs via voluntary human evaluation. We asked $10$ graduate students to evaluate $500$ \textit{<passage, question, answer>} samples: $200$ samples generated by the CS2S-VR-ACS model, $200$ samples generated by the GPT2-ACS model, and $100$ ground truth samples from the SQuAD 1.1 training dataset. All the samples are randomly shuffled, and each sample will be evaluated by 3 volunteers\footnote{The participation of the study was completely voluntary. The reviewers kindly offered their help as volunteers without being compensated in any form. There was no consequence for declining participating.}.
We collected responses of the following questionnaire:

\begin{itemize}
	\item \textbf{Is the question well-formed?} This is to check whether a given question is both grammatical and meaningful \cite{krishna2019generating}. Workers will select \textit{yes}, \textit{no}, or \textit{understandable}. The option \textit{understandable} is selected if a question is not totally  grammatically correct, but we can infer its meaning.
	\item \textbf{If the question is well-formed or understandable, is the question relevant to the passage?} Workers will select \textit{yes} if we can find the answer to the generated question in the passage.
	\item \textbf{If the question is relevant to the passage, is the answer actually a valid answer to the generated question?} Workers will select \textit{yes}, \textit{no} or \textit{partially}. The last option represents that the answer in our generated sample partially overlaps with the true answer in the passage.
\end{itemize}

Table~\ref{tab:quality} shows the evaluation results based on the sampled data.
First, we can see that even the ground truth samples from the SQuAD dataset are not totally well-formed, relevant or answered correctly. That is because we only use the sentence which contains the answer span as the context. However, about 20\% questions in SQuAD require paragraph-level context to be asked \cite{du2017learning}.
Second, $94\%$ of the questions generated by GPT2-ACS are well-formed or understandable, while the percentage is $71.5\%$ for the CS2S-VR-ACS model. We can see that although the BLEU scores of the GPT2-ACS model are lower than that of CS2S-VR-ACS, the results of GPT2-ACS are semantically better due to the knowledge learned from large-scale pre-training.
Third, we can see that most of the questions generated by both CS2S-VR-ACS and GPT2-ACS are relevant to the input passages.
Finally, most of the answers are also correct given the generated questions. This demonstrates the high quality of the question-answer pairs generated by our models.

\begin{figure}[tb]
\centering
\includegraphics[width=3.4in]{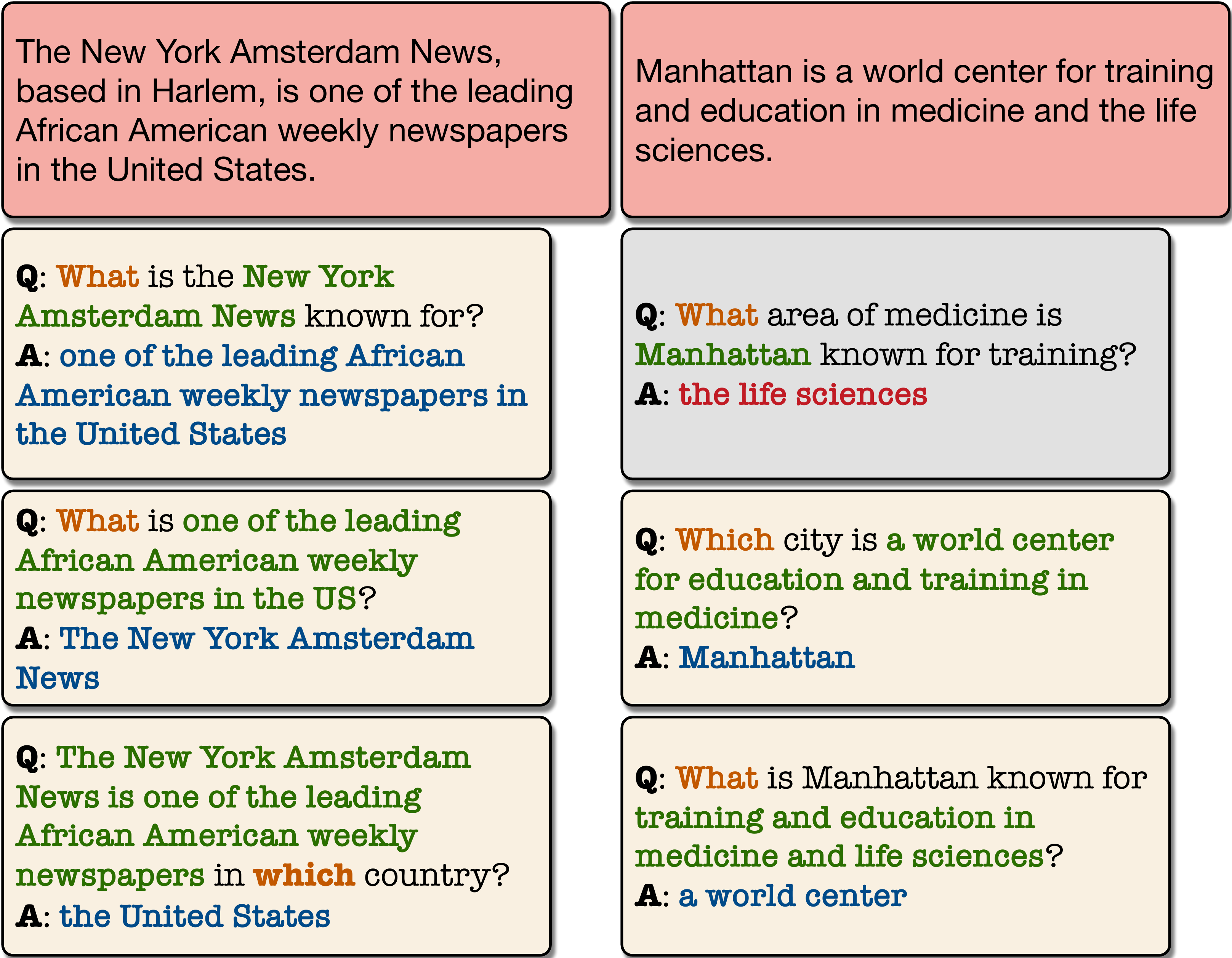}
\vspace{-5mm} 
\caption{Examples of the questions-answer pairs generated by our system.}
\label{fig:Showcase}
\vspace{-5mm} 
\end{figure}

Figure~\ref{fig:Showcase} is a running example to show the properties of generated questions. We can see that our ACS-aware question generation models are able to generate various questions based on different answers, question styles, and clue chunks. Compared with answer-aware question generation, our generation process is more controllable, and our generated questions are more diverse and of high quality. In some cases, the answer span does not match with the question.
We further performed pilot user studies to analyze the bad cases in our generated samples. For each question that is not well-formed, we ask workers to label the weakness of it.
The results of the study show that most of the errors are grammatically errors, type mismatches, meaningless, or incomplete information.
For CS2S-VR-ACS, about $56.7\%$ of the bad cases are grammatically incorrect; $29.2\%$ of them have the problem of type mismatch, e.g., the generated question starts with ``When'' when asking questions about a person; $14.2\%$ of them are grammatically correct, but meaningless; and $17.5\%$ of the questions do not express their meaning explicitly due to missing words or fuzzy pronouns. Similarly, for GPT2-ACS, the percentages of the above four problems are $40.4\%$ (grammatically incorrect), $46.2\%$ (type mismatches), $15.4\%$ (meaningless) and $11.6\%$ (incomplete or fuzzy information). Note that the sum of these percentages does not equal to 1. That is because each question may be labeled with multiple types of weaknesses. 

In order to reduce different errors and further improve the quality of generated questions, first, we need to incorporate the knowledge of natural language by methods such as large-scale pre-training. We can observe from Table~\ref{tab:quality} that most of the questions generated by GPT2-ACS are well-formed and grammatically correct. Second, we can utilize a better named entity recognition model to provide more accurate information about the named entity types of answer chunks. In this way, the type mismatch errors will be reduced. Last but not the least, the problem of semantic mismatching, meaningless, or information incompleteness can be reduced by training a better entailment model to enhance the data filter.

\begin{table}[tb]
\centering
\tabcolsep=0.11cm
\small
  \begin{tabular}{c|cc|cc}
    \toprule
    \multirow{2}{*}{\textbf{Experiments}} & \multicolumn{2}{c|}{\textbf{CS2S-VR-ACS}} & \multicolumn{2}{c}{\textbf{GPT2-ACS}} \\
    & \textbf{EM} & \textbf{F1}  & \textbf{EM} & \textbf{F1}  \\ 
    \hline
    SQuAD & 86.72 & 92.97 & 86.72 & 92.97  \\
    Generated & 71.14 & 83.53  & 74.47 & 85.64 \\
    Generated + SQuAD & 86.12 & 92.36 & 85.87 & 92.33 \\
    \bottomrule
  \end{tabular}
  \caption{Evaluating the question answering performance with different training datasets.}
  \label{tab:qa-exp}
  \vspace{-7mm}
\end{table}

\subsection{Applied to Question Answering}

We also perform quality test of the generated question-answer pairs by applying them to downstream machine reading comprehension tasks.
Specifically, we train different BERT-based question answering models based on the following settings:
\begin{itemize}
	\item \textbf{SQuAD}: in this experiment, we utilize the original SQuAD 1.1 training dataset to train a BERT-based QA model, and test the performance on the dev dataset. The performance is evaluated by exact match (EM) and F1-Score (F1) between predicted answer span and the true answer span \cite{rajpurkar2016squad}.
	\item \textbf{Generated}: in this experiment, we sample a training dataset from our generated questions, where the size is equal to the training dataset of SQuAD 1.1. Although our questions are generated from sentences, we utilize the paragraphs the sentences belong to as contexts when training QA models.
	\item \textbf{Generated + SQuAD}: in this experiment, we combine the original SQuAD training dataset with our generated training dataset to train the BERT-based QA model.
\end{itemize}
For all the QA experiments, the configurations are the same except the training datasets.

Table~\ref{tab:qa-exp} compares the performance of the resulting BERT-based QA models trained by above settings. Our implementation gives $86.72\%$ EM and $92.97\%$ F1 when trained on the original SQuAD dataset.
In comparison, the model trained by our generated dataset gives $74.47\%$ and $85.64\%$ F1.
When we combine the generated dataset with the SQuAD training set to train the QA model, the performance is not further improved.
The results are reasonable. First, the generated dataset contains noises which will influence the performance. Second, simply increasing the size of training dataset will not always help with improving the performance.
If most of the generated training samples are already answerable by the model trained over the original SQuAD dataset, they are not very helpful to further enhance the generalization ability of the model.

There are at least two methods to leverage the generated dataset to improve the performance of QA models.
First, we can utilize curriculum learning algorithms \cite{bengio2009curriculum} to select samples during training. We can select samples according to the current state of the model and the difficulties of the samples to further boost up the model's performance. Note that this requires us to remove the BERT-based QA model from our data filter, or set the threshold of filtering F1-score to be smaller.
Second, similar to \cite{gao2018difficulty}, we can further incorporate the difficulty information into our question generation models, and encourage the model to generate more difficult question-answer pairs.
We leave these to our future works.

Aside from machine reading comprehension, our system can be applied to many other applications. First, we can utilize it to generate exercises for educational purposes.
Second, we can utilize our system to  generate training datasets for a new domain by fine-tuning it with a small amount of labeled data from that domain. This will greatly reduce the human effort when we need to construct a dataset for a new domain. Last but not the least, our pipeline can be adapted  to similar tasks such as comment generation, query generation and so on.

\section{Related Work}
\label{sec:related}

In this section, we review related works on question generation.

\textbf{Rule-Based Question Generation}.
The rule-based approaches rely on well-designed rules or templates manually created by human to transform a piece of given text to questions \cite{heilman2010good,heilman2011automatic,chali2015towards}.
The major steps include preprocessing the given text to choose targets to ask about, and generate questions based on rules or templates \cite{sun2018answer}.
However, they require creating rules and templates by experts which is extremely expensive.
Also, rules and templates have a lack of diversity and are hard to generalize to different domains.

\textbf{Answer-Aware Question Generation}.
Neural question generation models are trained end-to-end and do not rely on hand-crafted rules or templates. The problem is usually formulated as answer-aware question generation, where the position of answer is provided as input.
Most of them take advantage of the encoder-decoder framework with attention mechanism \cite {serban2016generating,du2017learning,liu2019learning,zhou2017neural,song2018leveraging,hu2018aspect,du2018harvesting}.
Different approaches incorporate the answer information into generation model by different strategies, such as answer position indicator \cite{zhou2017neural,liu2019learning}, separated answer encoding \cite{kim2019improving}, embedding the relative distance between the context words and the answer \cite{sun2018answer} and so on. However, with context and answer information as input, the problem of question generation is still a one-to-many mapping problem, as we can ask different questions with the same input.

\textbf{Auxiliary-Information-Enhanced Question Generation}.
To improve the quality of generated questions, researchers try to feed the encoder with extra information. \cite{gao2018difficulty} aims to generate questions on different difficulty levels. It learns a difficulty estimator to get training data, and feeds difficulty as input into the generation model. \cite{krishna2019generating} learns to generate ``general'' or ``specific'' questions about a document, and they utilize templates and train classifier to get question type labels for existing datasets. \cite{hu2018aspect} identifies the content shared by a given question and answer pair as an aspect, and learns an aspect-based question generation model. \cite{gupta2019improving} incorporates knowledge base information to ask questions. Compared with these works, our work doesn't require extra labeling or training overhead to get the training dataset. Besides, our settings for question generation dramatically reduce the difficulty of the task, and achieve much better performance.

\textbf{Multi-task Question Generation}.
Another strategy is enhancing question generation models with correlated tasks. Joint training of question generation and answering models has improved the performance of individual tasks \cite{tang2017question,tang2018learning,wang2017joint,sachan2018self}.
\cite{liu2019learning} jointly predicts the words in input that is related to the aspect of the targeting output question and will be copied to the question. \cite{zhou2019type} predicts the question type based on the input answer and context. \cite{zhou2019multi} incorporates language modeling task to help question generation. \cite{zhang2019addressing} utilizes question paraphrasing and question answering tasks to regularize the QG model to generate semantically valid questions.

\section{Conclusion}
\label{sec:conclude}

In this paper, we propose ACS-aware question generation, a mechanism to generate question-answer pairs from unlabelled text corpus in a scalable way.  
By sampling meaningful tuples of clues, answers and question styles from the input text and use the sampled tuples to confine the way a question is asked, we have effectively converted the originally one-to-many question generation problem into a one-to-one mapping problem. We propose two neural network models for question generation from input passage given the selected clue, answer and question style, as well as discriminators to control the data generation quality.

We present extensive performance evaluation of the proposed system and models. Compared with existing answer-aware question generation models and models with auxiliary inputs or tasks, our ACS-aware QG model achieves significantly better performance, which confirms the importance of clue and style information.
We further resorted to voluntary human evaluation to assess the quality of generated data. Results show that our model is able to generate diverse and high-quality questions even from the same input sentence. Finally, we point out potential future directions to further improve the performance of our pipeline.

\clearpage

\bibliographystyle{ACM-Reference-Format}
\bibliography{main}

\end{document}